\documentclass{article}
\usepackage{ijcai26}

\usepackage{times}
\usepackage{url}

\usepackage[hidelinks,pdfa]{hyperref}

\usepackage[utf8]{inputenc}
\usepackage[small]{caption}
\usepackage{graphicx}
\usepackage{amsmath}
\usepackage{amsthm}
\usepackage{amssymb}
\usepackage{amsfonts}
\usepackage{booktabs}
\usepackage{multirow}
\usepackage{algorithm}
\usepackage{algorithmic}
\usepackage{enumitem}
\usepackage[switch]{lineno}

\hypersetup{
  pdftitle={A Foundation Model for Zero-Shot Logical Rule Induction},
  pdfauthor={Yin Jun Phua},
  pdfsubject={IJCAI 2026},
  pdfkeywords={inductive logic programming, zero-shot, rule induction, foundation model, neural-symbolic},
  pdflang={en-US}
}

\title{A Foundation Model for Zero-Shot Logical Rule Induction}

\author{
    Yin Jun Phua
    \affiliations
    Institute of Science Tokyo
    \emails
    phua@comp.isct.ac.jp
}

\begin{document}

\maketitle

\begin{abstract}
  Inductive Logic Programming (ILP) learns interpretable logical rules from data. Existing methods are transductive: their learned parameters are bound to specific predicates and require retraining for each new task. We introduce Neural Rule Inducer (NRI), a pretrained model for zero-shot rule induction. Rather than encoding literal identities, NRI represents literals using domain-agnostic statistical properties such as class-conditional rates, entropy, and co-occurrence, which generalize across variable identities and counts without retraining. The model consists of a statistical encoder and a parallel slot-based decoder. Parallel decoding preserves the permutation invariance of logical disjunction; an autoregressive decoder would instead impose an arbitrary clause order. Product T-norm relaxation makes rule execution differentiable, allowing end-to-end training on prediction accuracy alone. We evaluate NRI on rule recovery, robustness to label noise and spurious correlations, and zero-shot transfer to real-world benchmarks, and we believe this work opens up the possibility of foundation models for symbolic reasoning. Code and the reference checkpoint are available at \url{https://github.com/phuayj/neural-rule-inducer}. An extended version with full appendices is available at \url{https://arxiv.org/abs/2605.04916}.
\end{abstract}

\section{Introduction}

Inductive Logic Programming (ILP) systems learn interpretable logical rules purely from examples. For example, an ILP system analyzing patient records with symptoms and diagnoses might induce a rule: ``\textit{(fever $\land$ cough) $\lor$ (chills $\land$ body\_aches) $\to$ flu}.'' In high-stakes domains like healthcare and finance, black-box predictions are unacceptable. Transparent logical rules like this give us insights that we can actually trust and use.

Traditional symbolic ILP methods like \cite{Muggleton1994ILP,Quinlan1990FOIL,Srinivasan2001Aleph,inoue2014learning,Muggleton1995Progol,Cropper2021Popper} are sensitive to noise and often have to compromise between precision and coverage. Recently, differentiable approaches such as \cite{Evans2018LearningEQ,Gao2024DFORL,Fekri2025GLIDR} have been proposed. These methods have proven to be more robust to noise but they are still transductive, where learned weights are tied to specific predicates. A model trained on family relationships (\texttt{Parent}, \texttt{Grandparent}) cannot be transferred to biology (\texttt{Protein}, \texttt{Enzyme}). This means that for every new dataset, we need to retrain the models from scratch.

So how do we determine which variable (boolean attributes like \texttt{fever}, \texttt{cough}) is predictive of a label (targets like \texttt{flu})? We select variables by their statistical signatures rather than by name. Because these signatures are invariant to renaming and reordering and each literal is represented by a fixed-dimensional vector independent of $N$, they offer a plausible route to zero-shot transfer. A high class-conditional rate indicates a variable belongs in a rule, regardless of what it represents. In this paper we test this hypothesis by pretraining on synthetic DNFs and evaluating on held-out real-world tabular tasks without retraining.

This approach requires addressing three challenges. First, \textit{variable-sized problems}: generalization from fixed-size training data to problems with fewer or more variables. Second, \textit{inter-variable dependencies}: individual statistics miss redundancy and complementarity (e.g., XOR patterns). Third, \textit{synthetic-to-real transfer}: learning the abstract procedure of induction rather than overfitting to synthetic data.

\begin{figure*}[t]
  \centering
  \includegraphics[width=\textwidth]{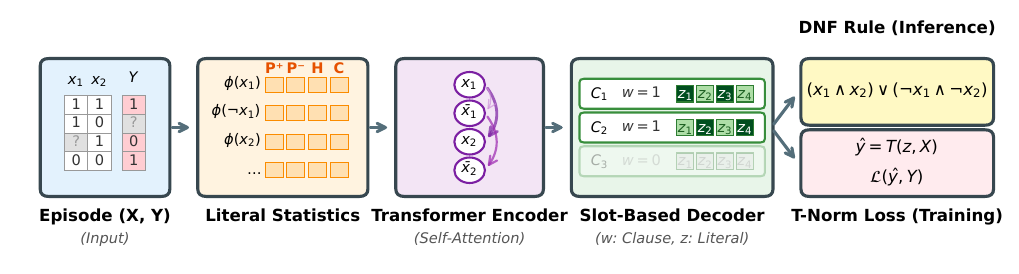}
  \caption{Neural Rule Inducer takes an episode $(X, Y)$ as input and calculates literal statistics. For each variable we calculate $\phi(x_i)$, $\phi(\neg x_i)$ which consists of class-conditional rates ($P^+$, $P^-$), entropy ($H$), and co-occurrence strength ($C$). We then apply cross-attention over these statistics. The slot-based decoder produces $K$ candidate clauses in parallel using learned literal gates $z$ and clause gates $w$. By evaluating the produced rule with T-norm, we can perform end-to-end training. The rules are discretized to then produce an interpretable DNF rule.}
  \label{fig:pipeline}
\end{figure*}

We present \textbf{Neural Rule Inducer (NRI)}, a foundation model~\cite{Bommasani2021FoundationModels} that consists of the following:
\begin{itemize}
  \item \textbf{Literal Statistics Encoder} encodes literals by statistical properties such as how often the literal is true and co-occurrences with other literals.
  \item \textbf{Parallel Slot-Based Decoder.} This module synthesizes multiple clauses in parallel using learned slot queries. Compared to autoregressive models, this module preserves the permutation invariance of logical disjunction.
  \item \textbf{T-Norm Training.} We use product t-norm relaxation to execute rules differentiably. This allows end-to-end training without explicit clause supervision.
  \item \textbf{Synthetic Data Training.} We train entirely on randomly generated boolean formulas. This trains the model to recognize the induction procedure itself rather than domain-specific patterns, and removes the limitation on training data size.
\end{itemize}
We show that a model trained entirely on synthetic boolean formulas can perform zero-shot rule induction on diverse real-world benchmarks. Figure~\ref{fig:pipeline} illustrates the overall architecture of NRI. While in this work, we focus on boolean variables, the statistical encoding framework can be extended to multi-valued and continuous domains via discretization or fuzzy predicates.

\section{Related Works}

\paragraph{Differentiable ILP.} \cite{Evans2018LearningEQ} proposed learning rules via gradient descent, but their method requires specifying rule templates and a fixed set of background predicates. NeuralLP \cite{Yang2017NeuralLP} and DRUM \cite{Sadeghian2019DRUM} extended this to knowledge base reasoning using TensorLog. Neural Theorem Provers~\cite{Rocktaschel2017NTP} introduced differentiable unification for end-to-end theorem proving. Logic Tensor Networks~\cite{Serafini2016LTN} use fuzzy semantics to integrate logical constraints into neural learning. DeepProbLog~\cite{Manhaeve2018DeepProbLog} extends probabilistic logic programming with neural predicates, enabling end-to-end learning of both neural and symbolic components. NeurASP~\cite{Yang2020NeurASP} integrates neural networks with answer set programming, allowing neural network outputs to serve as probabilistic inputs to logic programs. GLIDR \cite{Fekri2025GLIDR} generalized this to graph-like topologies, utilizing differentiable message passing to learn recursive and cyclic dependencies.

Most differentiable ILP systems instantiate parameters against a fixed predicate or schema inventory, so transferring to a new dataset typically requires retraining. Classical symbolic systems such as FOLD-R++~\cite{Wang2022FoldRPP}, which learns answer-set rules from mixed numerical and categorical data by top-down heuristic search, are similarly task-specific and are re-run from scratch on each new task.

Phua and Inoue \cite{Phua2024VAINN} proposed a model allowing zero-shot transfer learning. They addressed scaling issues by exploiting variable permutation symmetries. Our proposed method differs in mechanism where instead of utilizing the raw examples, we utilize statistical properties to handle missing and noisy data.

\paragraph{Generative Neuro-Symbolic AI.} LLM-based methods like ILP-CoT \cite{Muggleton2025HumanLike} and DeepSeek-Prover-V2 \cite{DeepSeek2025Prover} generate hypotheses by relying on knowledge encoded in language and human concepts. LINC~\cite{Olausson2023LINC} combines language models with first-order logic provers, using LLMs to translate natural language into formal logic for external verification. However, LLMs are not grounded in reality, they may use abstract symbols or notions that might not correspond to any measurable quantity in the real world. In contrast, our approach generates hypotheses directly from observed data and does not assume an explicit semantic model. TabPFN~\cite{Hollmann2023TabPFN} demonstrated that transformers trained on synthetic tabular data can perform in-context learning for classification. While TabPFN produces black-box predictions, our approach outputs interpretable logical rules.

\section{Background}

We want to learn a function $f: (\mathcal{X}, \mathcal{Y}) \to \mathcal{R}$ that takes input examples $\mathcal{X}$ (boolean variables) and labels $\mathcal{Y}$ and outputs a logical hypothesis $\mathcal{R}$ in Disjunctive Normal Form (DNF). Importantly, $f$ should work regardless of how many variables $N$ there are and what they represent. We should not need to retrain for each new problem.

\subsection{Disjunctive Normal Form (DNF)}

DNF is a disjunction of conjunctions: $\mathcal{R} = C_1 \lor C_2 \lor \dots \lor C_K$, where each clause $C_k$ is a conjunction of literals: $C_k = l_{k,1} \land l_{k,2} \land \dots$. A \textit{literal} $l_j$ is either a variable $x_i$ or its negation $\neg x_i$. DNF is a canonical form that can express any boolean function. Therefore, by learning DNFs we can learn any propositional rule.

\subsection{T-Norms}
A triangular norm (t-norm) is a mathematical operation on $[0,1]$ that relaxes logical conjunction to continuous values~\cite{Klement2000TNorms}. The product t-norm defines the following operations:
\begin{itemize}
    \item \textbf{Negation:} $\neg x = 1 - x$
    \item \textbf{Conjunction:} $x \land y = x \cdot y$
    \item \textbf{Disjunction:} $x \lor y = 1 - (1-x)(1-y)$ (via De Morgan)
\end{itemize}
Under product t-norm, the truth value of a clause $C_k$ containing literals indexed by set $S_k$ is:
\begin{equation}
    C_k = \prod_{i \in S_k} l_i
\end{equation}
where $l_i$ is the truth value of literal $i$. A DNF formula can then be calculated differentiably as follows $\mathcal{R} = 1 - \prod_k (1 - C_k)$.

\subsection{Inductive Logic Programming}
Inductive Logic Programming (ILP) systems learn logical rules from examples~\cite{Muggleton1994ILP}. Given positive examples $E^+$, negative examples $E^-$, and background knowledge $B$, an ILP system finds a hypothesis $H$ such that $B \land H \models E^+$ (completeness) and $B \land H \not\models E^-$ (consistency).

Two settings exist in the ILP literature. In \textit{learning from entailment}, examples are ground facts. The hypothesis must logically entail positives and not entail negatives. In \textit{learning from interpretations}~\cite{inoue2014learning}, each example is a complete state. Rules explain state transitions or classifications. Our setting mainly follows learning from interpretations, where each example is a complete boolean assignment. The output of our ILP system is a DNF rule that classifies all positive examples and none of the negative examples.

\subsection{Foundation Models}
A foundation model is a model trained on huge amounts of data and transfers the learning to downstream tasks without task-specific training~\cite{Bommasani2021FoundationModels}. Three properties define this paradigm: training on diverse data at scale, generalization beyond the training distribution, and adaptation to new tasks via prompting or fine-tuning. GPT and CLIP are examples in language and vision.

We apply this paradigm to rule induction. Our training data consists of millions of synthetic boolean formulas with diverse rule structures. The model does not learn weights for specific predicates. Instead, it learns to recognize statistical patterns of literals that allows the model to infer that the literal belongs to a rule. At inference, the model is able to induce rules for new domains without retraining.

\section{Neural Rule Inducer (NRI)}

In this section we propose NRI, an end-to-end differentiable framework that does domain-agnostic rule learning. Rather than learning weights for specific predicates for a specific domain, NRI learns to select variables based on their statistical properties within the domain.

\paragraph{Design Rationale.} NRI separates three roles: the statistical encoder provides identity-free cues about which literals matter, the example-conditioned encoder recovers which examples each literal covers, and the parallel decoder assembles clauses without imposing an artificial order on a disjunction. FiLM breaks symmetry among clause slots so different clauses can specialize. Section~\ref{sec:training_objective} and Table~\ref{tab:loss_ablation} study loss-level contributions; fuller architectural swaps are deferred because they would change these symmetry and transfer assumptions.

\subsection{Problem Formulation}

Our setup is structurally a meta-learning problem: the training distribution is over entire episodes, not individual literals. For a sampled $N$, the causal literal set is $\mathcal{L}_N = \{x_1, \neg x_1, \ldots, x_N, \neg x_N\}$; we sample a bounded DNF rule $R$ over $\mathcal{L}_N$, draw a causal matrix $X_{\text{c}} \in \{0,1\}^{M \times N}$, set $Y = R(X_{\text{c}})$, and then optionally add missingness, label noise, and concatenated spurious variables (Appendix~\ref{sec:synthetic_data_generation}). The hypothesis space at inference is therefore bounded DNFs over the literal set of the target task. At evaluation time we test on a fixed collection of UCI tasks rather than a parametric test distribution, with $\mathcal{L}_N$ determined by binarizing each task's features. Within a task, generalization from the support split to the held-out split is the usual i.i.d.\ setting; across tasks, transfer relies on the inductive-bias assumption that many binarized tasks admit sparse bounded-DNF descriptions. We do not provide formal cross-task guarantees here; the contribution is architectural and empirical.

Given $X \in \{0, 1\}^{M \times N}$ and $Y \in \{0, 1\}^M$ where $M$ is the number of examples and $N$ is the number of variables (or features), we would like to find a DNF hypothesis in the following form:
\begin{equation}
    \mathcal{R} = \bigvee_{k=1}^K \left( \bigwedge_j^L l_j \right)
    \label{eq:dnf}
\end{equation}
such that $\mathcal{R}(X) = Y$.

NRI can be separated into four stages (depicted in Figure~\ref{fig:pipeline}). First, we calculate statistical properties $\Phi \in \mathbb{R}^{2N \times D}$ for each variable. Then, we project these statistical properties via cross-attention over examples into $Z \in \mathbb{R}^{2N \times d}$. Next, we produce a DNF rule via slot-based attention. Finally, we evaluate DNF rule using differentiable T-norms for end-to-end training.

\subsection{Literal Statistics Encoder}
\label{sec:literal_stats}

To achieve zero-shot generalization with robustness to missing and noisy data, we encode each literal by its \textbf{statistical properties} over all $M$ examples. This differs from prior works that use raw samples directly as input.

For each literal $l_j$ (where $j \in \{1, \ldots, 2N\}$ indexes both positive literals $x_i$ and negations $\neg x_i$), we compute a feature vector $\phi_j \in \mathbb{R}^D$ containing:
\begin{equation}
    \phi_j = \left[ P(l_j | y{=}1), P(l_j | y{=}0), P(l_j), \mathcal{H}(l_j), \text{sgn}_j, \bar{c}_j, \dots \right]
    \label{eq:phi}
\end{equation}
where $P(l_j | y)$ denotes literal truth rates, $\mathcal{H}(l_j)$ is the binary entropy, $\text{sgn}_j \in \{0, 1\}$ indicates literal polarity (positive/negative), and $\bar{c}_j$ is the mean absolute co-occurrence with other literals. The full feature vector includes 18 components (see Appendix~\ref{sec:literal_features}).

These statistics are fed into an MLP to produce an initial embedding:
\begin{equation}
    h_j^{(0)} = \text{MLP}(\phi_j) \in \mathbb{R}^d
    \label{eq:h0}
\end{equation}

The observation-rate features in $\phi_j$ explicitly quantify how much of the input is observed. For example, if 30\% of the values for literal $j$ are missing among positive examples, the truth rate $P(l_j{=}1 \mid y{=}1)$ is computed only from the observed 70\%, and $\text{obs}^+ = 0.7$ indicates reduced statistical support. Noisy literals manifest similarly: truth rates move toward $0.5$ and entropy increases. As observation rates fall, the signal becomes correspondingly weaker because these statistics are estimated from fewer effective samples.

\subsection{Example-Conditioned Encoding}
\label{sec:example_attn}

The aggregate statistics $\phi_j$ are not claimed to be information-theoretically sufficient for arbitrary DNFs: two literals can share the same marginals yet cover different subsets of positive examples. The example-conditioned encoder is therefore used to restore this support-pattern information. Our claim is empirical adequacy of this compression rather than formal sufficiency.

Each example $m$ is represented by a key vector $e_m$ combining its label and literal values:
\begin{equation}
    e_m = \text{MLP}_y([y_m, 1{-}y_m, \mathbf{1}_{m}]) + \text{MLP}_x(l^{(m)})
    \label{eq:example_key}
\end{equation}
where $l^{(m)} \in [0,1]^{2N}$ is the vector of literal truth values for example $m$, and $\text{MLP}_x$ uses a 64-dimensional bottleneck.

\paragraph{Dynamic Dimension Adaptation.} $\text{MLP}_x$ is trained with a fixed input dimension $2N_{\text{train}}$. At inference, when facing problems with $N \neq N_{\text{train}}$, we adapt the linear layer on-the-fly. If $N < N_{\text{train}}$, we zero-pad the input to match the trained dimension. If $N > N_{\text{train}}$, we expand the first layer: trained weights are copied for the first $2N_{\text{train}}$ dimensions, and the remaining $2(N - N_{\text{train}})$ dimensions are initialized randomly. This adaptation affects only the auxiliary example-conditioned branch; the main literal-statistics pathway is already dimension-free in $N$. The 64-dimensional bottleneck limits the influence of newly initialized weights, so this mechanism is a pragmatic approximation rather than a principled invariance guarantee.

The literal embeddings are produced by applying attention to these example keys:
\begin{equation}
    h_j^{(1)} = h_j^{(0)} + \text{MultiHeadAttn}(h_j^{(0)}, \{e_m\}_{m=1}^M, \{e_m\}_{m=1}^M)
    \label{eq:h1}
\end{equation}

\subsection{Clause-Conditioned Encoding via FiLM}
\label{sec:film}

The embeddings $h_j^{(1)}$ are shared across all clause slots. Without differentiation, clause slots will tend to converge to identical patterns during training. We therefore apply Feature-wise Linear Modulation (FiLM)~\cite{Perez2018FiLM} to give each clause slot $k$ a unique view:
\begin{equation}
    h_{k,j} = \gamma_k \odot h_j^{(1)} + \beta_k
    \label{eq:film}
\end{equation}
where $\gamma_k, \beta_k \in \mathbb{R}^d$ are learnable per-clause parameters. We initialize $\beta_k$ orthogonally and $\gamma_k \sim \mathcal{N}(1, 0.5^2)$, which encourages clause slots to differentiate.

\subsection{Slot-Based Decoder}
\label{sec:decoder}

We treat the DNF synthesis problem as a \textbf{slot-filling problem}. Given $T$ clause slots, the model decides which literals to include in each clause and which clauses to activate. Unlike autoregressive generation, the design of parallel slots aims to preserve permutation invariance in the generated rule ($A \lor B \equiv B \lor A$).

The decoder is a 3-layer Transformer Decoder~\cite{Vaswani2017Transformer} with 4 attention heads. Each clause slot $k$ has a learnable query $q_k \in \mathbb{R}^d$. The decoder computes the following:
\begin{equation}
    s_k = \text{TransformerDec}(q_k + \bar{h}, \{h_{k,j}\}_{j=1}^{2N})
    \label{eq:clause_state}
\end{equation}
where $\bar{h} = \frac{1}{2N}\sum_j h_j^{(1)}$ is an average of all literal embeddings.

\subsubsection{Literal Gates (``AND'' Level)}
\label{sec:literal_gates}
For each clause $k$, we compute the probability a literal gets included:
\begin{equation}
    z_{k,j} = \sigma\left( \frac{1}{\sqrt{d}} \langle W_s s_k, W_h h_{k,j} \rangle + b \right)
    \label{eq:literal_gate}
\end{equation}
where $W_s, W_h \in \mathbb{R}^{d \times d}$ are learnable projections and $b$ is a bias term. Since a clause containing both $x_i$ and $\neg x_i$ is always false, we remove the literal with the lower score in each complementary pair.

\subsubsection{Clause Gates (``OR'' Level)}
\label{sec:clause_gates}
A clause gate $w_k \in [0, 1]$ determines whether slot $k$ is active:
\begin{equation}
    w_k = \sigma\left( \text{MLP}\left( \left[ \bar{h},\, \tilde{h}_k,\, s_k,\, p_k \right] \right) \right)
    \label{eq:clause_gate}
\end{equation}
where $\tilde{h}_k = \sum_j z_{k,j} h_{k,j} / \sum_j z_{k,j}$ is the clause's weighted literal summary, and $p_k = 1 - \prod_j (1 - z_{k,j})$ is the probability that at least one literal is selected (non-null probability).

\paragraph{Clause Selection at Inference.}\label{sec:clause_ranking} The deployed inference rule retains every clause whose gate satisfies $w_k \geq 0.5$. As a diagnostic for clause quality we additionally compute a discrimination score:
\begin{equation}
    d_k = \frac{1}{|Y^+|}\sum_{m: y_m=1} C_k^{(m)} - \frac{1}{|Y^-|}\sum_{m: y_m=0} C_k^{(m)}
    \label{eq:clause_rank}
\end{equation}
We explored top-$K'$ filtering by $d_k$; it improves exact-match rule recovery on synthetic benchmarks but did not help UCI accuracy in our experiments, so it is not used by default.

\subsection{Neuro-Symbolic Execution}
\label{sec:execution}

The decoder produces literal gates $z_{k,j}$ and clause gates $w_k$ that define a soft DNF rule. These gates are computed once per episode from all $M$ examples. We then evaluate this rule on each example $m$ using product t-norms.

For each clause $k$ and example $m$, the clause truth value is:
\begin{equation}
    C_k^{(m)} = \prod_{j=1}^{2N} \left( 1 - z_{k,j} \cdot (1 - l_j^{(m)}) \right)
    \label{eq:clause_truth}
\end{equation}
When $z_{k,j} \to 0$, the term becomes 1 (literal ignored). When $z_{k,j} \to 1$, the term reduces to $l_j^{(m)}$ (clause depends on literal $j$).

The final prediction is a combination of all clauses via the probabilistic OR:
\begin{equation}
    \hat{y}^{(m)} = 1 - \prod_{k=1}^K \left(1 - w_k \cdot C_k^{(m)}\right)
    \label{eq:prediction}
\end{equation}
The prediction is high when at least one active clause ($w_k \approx 1$) is satisfied ($C_k^{(m)} \approx 1$). The entire computation from Eq.~\ref{eq:phi} to Eq.~\ref{eq:prediction} is differentiable, enabling end-to-end training.

\subsection{Training Strategy}

The combinatorial search space for ILP problems makes cold-start training difficult, particularly in a foundation model setting. To overcome this, we employ these techniques:
\begin{enumerate}
    \item \textbf{Synthetic Pre-training:} We train the model on random boolean formulas, allowing it to learn the induction algorithm without real-world data.
    \item \textbf{Spurious Environment Training:} We inject spurious features into each episode with \textit{opposite} correlations across two environments (first and second half of examples). After shuffling examples, these features should be \textit{marginally independent} from the label. The model must then learn to identify features that perform well on one environment but poorly on the other. See Appendix~\ref{sec:spurious_env} for details.
    \item \textbf{Clause Dropout:} During training, we randomly drop 25\% of clause slots (keeping at least 2) to prevent any single clause from dominating.
\end{enumerate}

\subsection{Training Objective}
\label{sec:training_objective}

We utilize a multi-objective loss function to balance between accuracy, slot utilization, clause diversity, gate sharpness, margin enforcement, and counterfactual necessity:
\begin{equation}
    \mathcal{L} = \mathcal{L}_{\text{cov}} + \lambda_b \mathcal{L}_{\text{bal}} + \lambda_r \mathcal{L}_{\text{rep}} + \lambda_e \mathcal{L}_{\text{ent}} + \lambda_m \mathcal{L}_{\text{mm}} + \lambda_{\text{cf}} \mathcal{L}_{\text{cf}}
\end{equation}

The difficulty in training a stable, parallel slot decoder with logical consideration necessitates such a complex objective. We describe each loss component in detail in the following paragraphs.

\paragraph{Coverage Loss.} Binary cross-entropy between predictions $\hat{y}$ and labels $y$.

\paragraph{Clause Slot Load-Balancing.} Without explicit regularization, the T-norm calculation tends to lead to a few clause slots dominating while others receive vanishing gradients. To overcome this, we employ multiple complementary balancing losses from the Mixture-of-Experts literature~\cite{Fedus2022SwitchTransformers}. First, we use the auxiliary load-balancing loss from Switch Transformers: $K \cdot \sum_k u_k \cdot f_k$, where $u_k$ is the mean routing probability to slot $k$ and $f_k$ is the fraction of assignments. Second, we apply a CV$^2$ (coefficient of variation squared) loss on normalized clause slot usage. Let $\bar{w}_k = \frac{1}{B}\sum_{b} w_k^{(b)}$ denote the mean clause gate activation across a batch:
\begin{equation}
    \mathcal{L}_{\text{bal}} = K \cdot \sum_{k=1}^{K} \left( \frac{\bar{w}_k}{\frac{1}{K}\sum_{k'} \bar{w}_{k'}} - 1 \right)^2
\end{equation}
We additionally apply CV$^2$ on raw clause gate activations (before normalization) to prevent clause slots from going permanently inactive. These losses produce gradients that counteract the task loss gradients, reducing clause slot utilization variance from 0.35 to 0.003 in our experiments.

\paragraph{Max Margin Coverage.} BCE rewards spreading probability across clauses (diffusion), while clause gates reward specialization. In our experiments, this conflict caused clause gate entropy to oscillate between low values (few clauses active) and high values (many clauses active).

We add a max-margin loss~\cite{Cortes1995SVM} that only penalizes the \textit{best} clause when it falls short of a margin. This removes the incentive for diffusion:
\begin{equation}
    \begin{aligned}
        \mathcal{L}_{\text{mm}} = {} & \mathbb{E}_{y=1}[\max(0, \tau^+ - C_{\max})] \\
                                     & + \mathbb{E}_{y=0}[\max(0, C_{\max} - \tau^-)]
    \end{aligned}
\end{equation}
where $C_{\max} = \max_k C_k$ is the maximum clause truth value, and $\tau^+ = 0.7$, $\tau^- = 0.3$. Unlike BCE which pushes all clauses to cover examples, max-margin allows multiple clause slots to cover the same pattern. This stabilizes training at low entropy ($\sim$0.26) where clauses use generalizable features rather than differentiating via rare attributes.

\paragraph{Counterfactual Necessity.} BCE alone cannot distinguish necessary literals from spurious ones. A literal that only happens to correlate with the label may be selected even without causal relationship. To address this, we add a counterfactual test: if a literal is selected it should be causal, i.e. flipping its value should break the clause. We define the counterfactual loss as $\mathcal{L}_{\text{cf}} = \mathcal{L}_{\text{nec}} + \mathcal{L}_{\text{spur}} + \lambda_o \mathcal{L}_{\text{ovl}} + \lambda_c \mathcal{L}_{\text{cf-bal}}$, where $\mathcal{L}_{\text{ovl}}$ ($\lambda_o{=}0.1$) penalizes pairs of clauses that both cover the same positive example and $\mathcal{L}_{\text{cf-bal}}$ ($\lambda_c{=}0.01$) is a mild regularizer encouraging balanced clause usage on positives within the CF objective; full forms are in Appendix~\ref{sec:cf_extra_terms}.

\textbf{Necessity:} BCE tends to reward correct predictions regardless of whether selected literals are causally necessary or merely correlated. So if a literal that has no causal relationship happens to be included in a correct prediction, the model might learn a spurious correlation. To overcome this, we use this loss to provide gradient signal toward causal selection. For positive examples, we flip selected literals (gates $z_{s,j}$ above threshold) and recompute clause truth $C_k'$. The loss minimizes post-flip clause truth:
\begin{equation}
    \mathcal{L}_{\text{nec}} = \mathbb{E}_{y=1}\left[\sum_k r_k \cdot C_k'\right]
\end{equation}
where $r_k = \text{softmax}_k(C_k)$ is a weight applied to each clause to emphasize important clauses.

\textbf{Spuriousness:} On the other hand, if a model decided that a literal should be ignored (gate $z_{s,j}$ below threshold), the model claims it is not part of the rule. If the prediction changes when we flip an ignored literal, the model was implicitly relying on something it claimed to ignore. This loss penalizes such inconsistency. For positive examples, we flip ignored literals and penalize if the prediction changed due to the flip:
\begin{equation}
    \mathcal{L}_{\text{spur}} = \mathbb{E}_{y=1}\left[|\hat{y} - \hat{y}'|\right]
\end{equation}
where $\hat{y}'$ is the result of the T-norm after flipping ignored literals.

\section{Experiments}

We train exclusively on synthetic boolean expressions and test zero-shot on real-world UCI datasets~\cite{Dua2017UCI} and benchmarks probing rule complexity, noise robustness, and spurious variable robustness. For each synthetic episode we sample $N \sim \mathrm{Unif}\{6, \ldots, 12\}$ and $M \sim \mathrm{Unif}\{24, \ldots, 48\}$ independently, and DNF rules with up to $K=6$ clauses and $L=4$ literals per clause. Each episode includes 3 spurious environment features (Appendix~\ref{sec:synthetic_data_generation}). The decoder uses $T=8$ clause slots with 25\% dropout. Training uses AdamW (lr=$6 \times 10^{-4}$, weight decay $10^{-2}$), batch size 8192, and 500 steps.

\subsection{Baseline Context}

Table~\ref{tab:exp8b_baseline_comparison} compares NRI against 8 baselines on 14 UCI datasets. Continuous features are binarized using median thresholding (e.g., $\texttt{age}_{>\text{med}}$ is 1 if age exceeds the median). Missing values are preserved: NRI handles them through observation-rate features and by treating unknown literals as 0.5 (maximally uncertain) during rule evaluation. Baselines use median imputation.

NRI is evaluated \textit{zero-shot} whereas other baseline methods are trained per dataset. Notably, 12 of 14 datasets have more features than the training range ($N > 12$), testing out-of-distribution generalization to larger schemas. All methods are evaluated under 5-fold stratified cross-validation: in each fold, one fold ($\sim$20\% of the data) acts as the support set on which the rule is induced, and accuracy is reported on the remaining $\sim$80\%. This support-set size targets the low-data regime where zero-shot transfer is most valuable. We compare against gradient boosting methods (XGBoost, LightGBM), generalized additive models (EBM), classical rule learners (RIPPER, RuleFit, FIGS), decision trees (DT), and neural DNF (N-DNF). Full descriptions of the baselines are in Appendix~\ref{sec:baseline_descriptions}.


\begin{table*}[t]
\centering
\scriptsize
\setlength{\tabcolsep}{3pt}
\begin{tabular*}{\textwidth}{@{\extracolsep{\fill}}lrrrrrrrrr|rr@{}}
\toprule
Dataset & $N$ & XGB & LGBM & EBM & RIPPER & RuleFit & FIGS & DT & N-DNF & NRI$^\dagger$ & Gap \\
\midrule
adult & 105 & 82.2 $\pm$ 0.2 & 82.7 $\pm$ 0.2 & \textbf{83.8 $\pm$ 0.1} & 81.9 $\pm$ 0.3 & 82.6 $\pm$ 0.2 & 82.8 $\pm$ 0.2 & 82.1 $\pm$ 0.2 & 83.0 $\pm$ 0.1 & 69.6 $\pm$ 4.4 & -14.2 \\
breast-cancer & 9 & 92.0 $\pm$ 1.4 & 65.5 $\pm$ 0.0 & \textbf{93.0 $\pm$ 1.2} & 88.3 $\pm$ 1.4 & 90.8 $\pm$ 2.2 & 89.6 $\pm$ 2.3 & 89.3 $\pm$ 2.6 & 90.3 $\pm$ 1.3 & 88.3 $\pm$ 0.3 & -4.7 \\
car & 21 & 80.5 $\pm$ 1.1 & 75.0 $\pm$ 1.6 & 81.2 $\pm$ 0.3 & 90.3 $\pm$ 0.9 & \textbf{94.2 $\pm$ 0.6} & 91.8 $\pm$ 1.3 & 78.7 $\pm$ 1.3 & 93.3 $\pm$ 1.0 & 51.2 $\pm$ 5.4 & -43.0 \\
credit & 46 & 82.1 $\pm$ 1.4 & 55.5 $\pm$ 0.0 & 82.2 $\pm$ 1.6 & \textbf{83.1 $\pm$ 2.1} & 81.7 $\pm$ 1.8 & 81.2 $\pm$ 1.7 & 81.3 $\pm$ 1.7 & 81.4 $\pm$ 2.6 & 71.5 $\pm$ 7.3 & -11.6 \\
diabetes & 8 & 67.4 $\pm$ 2.6 & 65.1 $\pm$ 0.0 & 66.9 $\pm$ 1.8 & 66.8 $\pm$ 2.2 & 65.5 $\pm$ 1.9 & 64.9 $\pm$ 2.0 & 65.3 $\pm$ 1.6 & 64.4 $\pm$ 2.3 & \textbf{68.0 $\pm$ 2.4} & +0.0 \\
german & 61 & 66.9 $\pm$ 1.1 & \textbf{70.0 $\pm$ 0.0} & 68.1 $\pm$ 1.7 & 58.3 $\pm$ 3.2 & 65.8 $\pm$ 1.0 & 64.6 $\pm$ 2.0 & 64.6 $\pm$ 2.4 & 68.1 $\pm$ 1.4 & 59.8 $\pm$ 3.8 & -10.2 \\
hepatitis & 32 & 79.4 $\pm$ 0.0 & 79.4 $\pm$ 0.0 & \textbf{81.2 $\pm$ 2.0} & 20.6 $\pm$ 0.0 & 20.6 $\pm$ 0.0 & 72.1 $\pm$ 3.2 & 73.8 $\pm$ 4.3 & 68.5 $\pm$ 5.3 & 55.9 $\pm$ 3.7 & -25.3 \\
ionosphere & 34 & 66.3 $\pm$ 4.3 & 64.1 $\pm$ 0.0 & 69.1 $\pm$ 3.9 & 59.5 $\pm$ 3.3 & \textbf{69.2 $\pm$ 6.0} & 63.8 $\pm$ 6.3 & 65.4 $\pm$ 5.1 & 67.8 $\pm$ 4.5 & 62.8 $\pm$ 3.9 & -6.4 \\
kr-vs-kp & 73 & 93.2 $\pm$ 0.6 & 90.7 $\pm$ 1.9 & 92.5 $\pm$ 0.5 & 89.9 $\pm$ 1.2 & \textbf{93.8 $\pm$ 0.7} & 92.7 $\pm$ 0.8 & 92.4 $\pm$ 0.6 & 92.0 $\pm$ 0.8 & 72.3 $\pm$ 5.3 & -21.5 \\
mushroom & 116 & 99.0 $\pm$ 0.2 & 99.1 $\pm$ 0.2 & 99.4 $\pm$ 0.1 & 98.5 $\pm$ 0.4 & \textbf{99.5 $\pm$ 0.2} & 99.4 $\pm$ 0.2 & 99.3 $\pm$ 0.2 & 98.8 $\pm$ 0.3 & 87.8 $\pm$ 3.9 & -11.7 \\
nursery & 27 & 90.5 $\pm$ 0.5 & \textbf{91.2 $\pm$ 0.4} & 91.1 $\pm$ 0.7 & 86.2 $\pm$ 1.0 & 88.5 $\pm$ 0.8 & 87.1 $\pm$ 0.5 & 84.8 $\pm$ 0.7 & 86.6 $\pm$ 0.7 & 71.3 $\pm$ 4.3 & -19.9 \\
spambase & 57 & 89.7 $\pm$ 0.6 & 89.8 $\pm$ 0.4 & \textbf{91.1 $\pm$ 0.6} & 85.7 $\pm$ 1.2 & 89.8 $\pm$ 0.4 & 85.7 $\pm$ 0.7 & 85.0 $\pm$ 1.0 & 87.8 $\pm$ 0.7 & 71.9 $\pm$ 0.7 & -19.2 \\
tic-tac-toe & 27 & 69.4 $\pm$ 2.6 & 65.3 $\pm$ 0.0 & \textbf{70.2 $\pm$ 1.2} & 56.0 $\pm$ 4.4 & 65.7 $\pm$ 1.5 & 64.8 $\pm$ 2.0 & 64.8 $\pm$ 1.3 & 67.8 $\pm$ 2.7 & 56.6 $\pm$ 2.5 & -13.6 \\
vote & 32 & \textbf{92.9 $\pm$ 1.8} & 61.4 $\pm$ 0.0 & 87.4 $\pm$ 2.1 & 91.6 $\pm$ 2.2 & 91.1 $\pm$ 1.6 & 90.5 $\pm$ 1.7 & 90.7 $\pm$ 2.4 & 91.8 $\pm$ 1.6 & 88.3 $\pm$ 1.8 & -4.6 \\
\midrule
\textbf{Mean} &  & 82.2 $\pm$ 11.2 & 75.4 $\pm$ 13.5 & \textbf{82.7 $\pm$ 10.6} & 75.5 $\pm$ 21.0 & 78.5 $\pm$ 20.4 & 80.8 $\pm$ 12.4 & 79.8 $\pm$ 11.5 & 81.6 $\pm$ 11.8 & 69.7 $\pm$ 12.0 & -13.0 \\
\bottomrule
\end{tabular*}
\caption{Baseline comparison on 14 UCI datasets (5-fold CV accuracy \%). $N$ = number of boolean features after median binarization. NRI is trained on $N \in [6, 12]$; 12/14 datasets are OOD ($N > 12$). Methods: gradient boosting (XGB, LGBM), GAM (EBM), rule-based (RIPPER, RuleFit, FIGS, DT), neural DNF (N-DNF), and our zero-shot NRI$^\dagger$. Best per row in bold.}
\label{tab:exp8b_baseline_comparison}
\end{table*}

Zero-shot NRI achieves 69.7\%, 13 points below EBM, which is unsurprising because EBM is trained separately on each dataset. Performance is strongest on the two in-distribution datasets (diabetes with $N{=}8$, breast-cancer with $N{=}9$), where NRI achieves 68.0\% and 88.3\% respectively.

Performance varies substantially across datasets due to two factors evident in the experimental data. Firstly, we converted car and nursery which are originally multi-class problems (4 and 5 classes respectively) to single class, the dataset is difficult to classify with just simple DNF rules. Second, kr-vs-kp requires all 8 clause slots for classification, which confirms that some datasets genuinely require more clauses than NRI's training distribution ($K{\leq}6$).

The learned rules are interpretable. For diabetes prediction (68.0\% accuracy), NRI induces: $(\texttt{glucose}_{>\text{med}} \land \texttt{age}_{>\text{med}})$, capturing that elevated plasma glucose combined with older age predicts diabetes. For breast cancer diagnosis (88.3\% accuracy): $(\texttt{cell\_size}_{>\text{med}} \land \texttt{cell\_shape}_{>\text{med}} \land \texttt{bare\_nuclei}_{>\text{med}})$, identifying that larger, irregularly shaped cells with prominent nuclei indicate malignancy. These are clinically plausible summaries of the binarized benchmarks, not validated medical decision rules, but they show NRI can recover human-readable patterns without task-specific training.

\subsection{Rule Complexity Scaling}

We evaluate rule recovery on synthetic DNF formulas with varying clause count $K \in \{1,2,3,4\}$ and literals per clause $L \in \{1,2,3\}$, testing at $N{=}12$ features (in-distribution, matching training range $N \in [6,12]$). For each $(K, L)$ configuration, we generate 200 random DNF rules per seed across 10 seeds and measure logical equivalence between the predicted and ground-truth rules.

\begin{figure}[t]
    \centering
    \includegraphics[width=0.9\columnwidth]{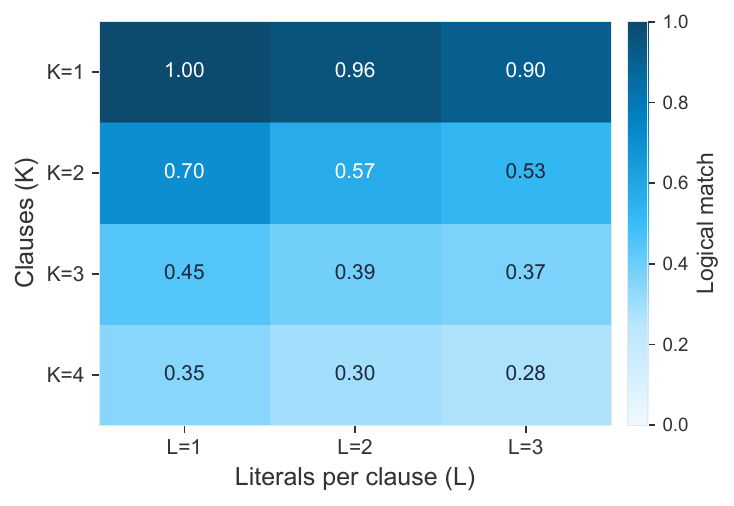}
    \caption{Heatmap of logical match rate (\%) across rule complexity dimensions at $N{=}12$. Darker colors indicate higher accuracy.}
    \label{fig:exp1a_heatmap}
\end{figure}

Figure~\ref{fig:exp1a_heatmap} shows that recovery degrades with complexity along both dimensions. For the simplest rules ($K{=}1$, $L{=}1$), logical match reaches 99.5\%. This drops to 91.5\% at $L{=}2$ and 81.7\% at $L{=}3$. Increasing clause count has a larger effect: at $K{=}4$, $L{=}1$, recovery falls to 34.2\%, and the most complex rules ($K{=}4$, $L{=}3$) achieve 24.0\% logical match. Prediction accuracy remains high (85--100\%) even when logical match is lower. The clause dimension ($K$) dominates difficulty because multi-clause rules require discovering multiple independent patterns, while longer clauses ($L$) only demand more precise literal selection within a single pattern.

\subsection{Label Noise Robustness}

\begin{figure}[t]
    \centering
    \includegraphics[width=0.9\columnwidth]{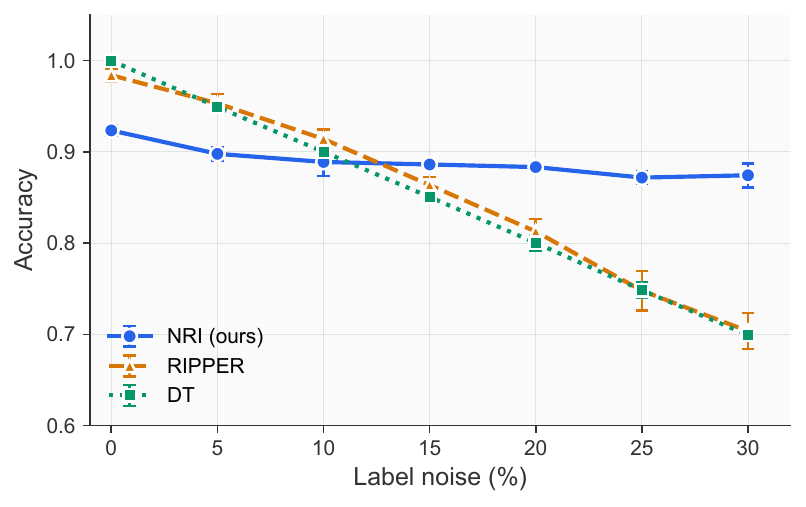}
    \caption{Noise robustness comparison. NRI (blue) maintains stable accuracy as noise increases. RIPPER (orange) and DT (green) degrade sharply. The crossover occurs around 15\% noise.}
    \label{fig:exp9a_noise}
\end{figure}

We evaluate robustness to label noise (random flips) against two interpretable baselines, RIPPER and DT. Figure~\ref{fig:exp9a_noise} shows NRI accuracy stays stable (92.3\%$\to$87.4\% at 30\% noise), while RIPPER and DT degrade sharply (98.4\%$\to$70.3\% and 100\%$\to$69.9\% respectively). Symbolic methods are more accurate at low noise by fitting training data exactly, but beyond 15\% noise NRI's statistical encoding provides implicit regularization and outperforms both baselines by 17 percentage points at 30\% noise.

\subsection{Spurious Variable Robustness}

\begin{figure}[t]
    \centering
    \includegraphics[width=0.9\columnwidth]{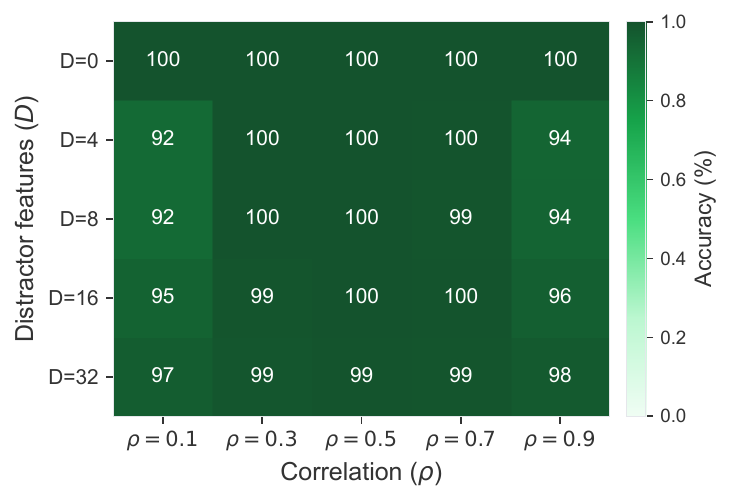}
    \caption{Accuracy heatmap across distractor count ($D$) and correlation strength ($\rho$). Accuracy remains above 92\% even with many highly-correlated spurious variables.}
    \label{fig:exp9b_spurious}
\end{figure}

We test whether NRI ignores spurious features that correlate with labels but are not in the true rule. We append $D \in \{0, 4, 8, 16, 32\}$ distractors per episode with $P(d{=}1 | Y{=}1) = \rho$ and $P(d{=}1 | Y{=}0) = 1{-}\rho$ for $\rho \in \{0.1, \ldots, 0.9\}$, statistically predictive but causally irrelevant features. Figure~\ref{fig:exp9b_spurious} shows accuracy remains above 92\% across all settings (97.6\% at $D{=}32, \rho{=}0.9$; 96.7\% at $D{=}32, \rho{=}0.1$ as the model learns to ignore the noise).

\subsection{Computational Scaling}
\label{sec:computational_scaling}

We measure inference time and memory as problem size ($N$) and example count ($M$) vary. Latency stays nearly constant ($\sim$7.5ms) as $M$ grows from 32 to 512, and increases sub-linearly with $N$ (4.2ms$\to$11.8ms for a 32$\times$ increase from $N{=}16$ to $N{=}512$); memory scales $O(N^2)$ due to attention. At $N{=}512$, inference completes in under 12ms with 593MB peak memory. Full curves are in Appendix~\ref{sec:scaling_app}.

\subsection{Loss Ablation Study}

\begin{table}[t]
\centering
\resizebox{\columnwidth}{!}{%
\begin{tabular}{lcccc}
\toprule
Ablation & UCI Accuracy (\%) & $\Delta$ Acc & Avg Clauses & Avg Literals \\
\midrule
Full (baseline) & 74.8 $\pm$ 3.8 & -- & 7.3 $\pm$ 1.2 & 3.2 $\pm$ 0.2 \\
-- CF Necessity & 74.1 $\pm$ 4.7 & -0.6 & 7.8 $\pm$ 0.5 & 3.1 $\pm$ 0.3 \\
-- Max-Margin & 72.0 $\pm$ 2.4 & -2.8 & 7.7 $\pm$ 0.4 & 3.2 $\pm$ 0.2 \\
-- Slot Balance & 73.3 $\pm$ 3.7 & -1.5 & 7.0 $\pm$ 1.5 & 3.0 $\pm$ 0.5 \\
\bottomrule
\end{tabular}}
\caption{Loss function ablation study. Each row removes one loss component.}
\label{tab:loss_ablation}
\end{table}

Each loss contributes (Table~\ref{tab:loss_ablation}): removing CF Necessity, Max-Margin Coverage, or Slot Balance drops UCI accuracy by 0.7\%, 2.8\%, and 1.5\% respectively, with characteristic failure modes (more spurious-literal selection, unstable gate entropy, clause monopoly with slot variance jumping from 0.003 to 0.35). Rule complexity (7--8 clauses, 3 literals) is similar across ablations, so these losses primarily affect \textit{which} literals are selected. The full model achieves the highest accuracy (74.8\%) while maintaining interpretable rule sizes.

\section{Conclusion}

We presented Neural Rule Inducer (NRI), a foundation model for zero-shot rule induction trained entirely on synthetic Boolean formulas. By encoding literals through identity-free statistical properties (class-conditional rates, entropy, co-occurrence) and synthesizing clauses with a parallel slot-based decoder under product T-norm semantics, NRI is trained end-to-end on prediction accuracy alone and induces interpretable DNF rules on new domains without retraining. We believe this work opens up the possibility of foundation models for symbolic reasoning. Future work extends to multi-valued and continuous variables and to first-order logic with relational predicates.

\section*{Acknowledgements}

This work was supported by JSPS KAKENHI Grant Number 25K21269. This study was carried out using the TSUBAME4.0 supercomputer at Institute of Science Tokyo. The author also thanks Professor Katsumi Inoue for a helpful discussion on this research.

\bibliographystyle{named}
\bibliography{paper}

\appendix

\section{Synthetic Data Generation}
\label{sec:synthetic_data_generation}

Training a foundation model for rule induction requires diverse examples spanning the space of possible logical rules. Since real-world datasets are limited in quantity and coverage, we train exclusively on synthetically generated episodes. This section describes the data generation procedure in detail.

\subsection{Episode Structure}

Each training episode $\mathcal{E} = (X, Y, \mathcal{R}^*)$ consists of:
\begin{itemize}
    \item $X \in \{0, 1\}^{M \times (N+S)}$: Boolean feature matrix with $M$ examples, $N$ causal variables, and $S$ spurious variables
    \item $Y \in \{0, 1\}^M$: Binary labels
    \item $\mathcal{R}^*$: Ground-truth DNF rule defined over variables $\{1, \dots, N\}$
\end{itemize}

The number of causal variables $N$ and examples $M$ are sampled from discrete uniform distributions for each episode: $N \sim \text{Unif}\{N_{\min}, \dots, N_{\max}\}$ and $M \sim \text{Unif}\{M_{\min}, \dots, M_{\max}\}$. In our experiments, $N \in \{6, \dots, 12\}$ and $M \in \{24, \dots, 48\}$.

\subsection{DNF Rule Sampling}

For each episode, we sample a random DNF rule $\mathcal{R}^* = C_1 \lor C_2 \lor \dots \lor C_K$ where each clause $C_k \subseteq \{1, \dots, N\} \times \{+, -\}$ is a set of signed literals. The sampling procedure is:

\begin{enumerate}
    \item Sample the number of clauses $K \sim \text{Unif}\{1, \dots, K_{\max}\}$
    \item For each clause $C_k$:
    \begin{enumerate}
        \item Sample the clause length $L_k \sim \text{Unif}\{1, \dots, \min(L_{\max}, N)\}$
        \item Sample $L_k$ distinct variable indices without replacement from $\{1, \dots, N\}$
        \item For each selected variable $i$, sample polarity $p_i \sim \text{Bernoulli}(0.5)$ to form literal $(i, p_i)$
    \end{enumerate}
\end{enumerate}

This procedure ensures that clauses contain no duplicate variables and that literal polarities are balanced. We do not canonicalize rules (e.g., by removing subsumed clauses), accepting some redundancy in exchange for sampling simplicity. In our experiments, $K_{\max} = 6$ and $L_{\max} = 4$.

\subsection{Example Generation}

Given the sampled rule $\mathcal{R}^*$, we generate the causal feature matrix $X_{\text{causal}} \in \{0,1\}^{M \times N}$ and labels $Y$:

\begin{enumerate}
    \item Sample each entry $X_{m,n} \sim \text{Bernoulli}(0.5)$ independently
    \item Compute labels by evaluating the DNF rule:
    \begin{equation}
        Y_m = \mathcal{R}^*(X_m) = \bigvee_{k=1}^K \bigwedge_{(i, p) \in C_k} \ell_p(X_{m,i})
    \end{equation}
    where $\ell_+(x) = x$ and $\ell_-(x) = 1 - x$
\end{enumerate}

The uniform random sampling produces class imbalance that depends on rule structure. A single clause of length $L$ yields $P(Y{=}1) = 2^{-L}$, so longer clauses produce fewer positives. Adding more clauses increases the positive rate via union, partially offset by clause overlap. This creates diverse class ratios across episodes.

\subsection{Spurious Environment Features}
\label{sec:spurious_env}

A key challenge in rule induction is distinguishing \textit{causal} features (those in the true rule) from \textit{spurious} features (correlated with the label but not part of the rule). We augment each episode with $S$ spurious variables that exhibit \textit{environment-dependent} correlations.

The examples are conceptually split into two environments: Environment 1 (examples $1$ to $\lfloor M/2 \rfloor$) and Environment 2 (examples $\lfloor M/2 \rfloor + 1$ to $M$). For each spurious feature $s$, we sample values with opposite correlations across environments. Let $\rho \in (0, 0.5)$ be the flip rate parameter:

\begin{itemize}
    \item \textbf{Environment 1:} $P(s{=}1 | Y{=}1) = \rho$, $P(s{=}1 | Y{=}0) = 1 - \rho$
    \item \textbf{Environment 2:} $P(s{=}1 | Y{=}1) = 1 - \rho$, $P(s{=}1 | Y{=}0) = \rho$
\end{itemize}

After generating spurious features, we concatenate them to form $X = [X_{\text{causal}} \,|\, X_{\text{spurious}}] \in \{0,1\}^{M \times (N+S)}$, then randomly permute all examples to hide environment boundaries.

\paragraph{Marginal Independence by Design.} An important property of this construction is that, marginally over the full dataset, each spurious feature is \textit{independent} of the label:
\begin{equation}
    P(s{=}1 | Y{=}1) = \tfrac{1}{2}\rho + \tfrac{1}{2}(1-\rho) = \tfrac{1}{2} = P(s{=}1 | Y{=}0)
\end{equation}
This means simple correlation-based selection cannot distinguish spurious from causal features. However, spurious features exhibit a characteristic pattern: they ``work'' for half the examples and ``fail'' for the other half. The model must learn to detect this inconsistency through the statistical encoder's attention over examples, combined with the counterfactual necessity loss that penalizes selecting features whose removal does not change predictions.

In our experiments, we use $S = 3$ spurious features with flip rate $\rho = 0.3$.

\subsection{Data Generation Parameters}

Table~\ref{tab:synthetic_params} summarizes the data generation parameters used in our experiments.

\begin{table}[t]
\centering
\small
\begin{tabular}{ll}
\toprule
\textbf{Parameter} & \textbf{Value} \\
\midrule
Causal variables ($N$) & $\{6, \dots, 12\}$ \\
Examples per episode ($M$) & $\{24, \dots, 48\}$ \\
Max clauses ($K_{\max}$) & 6 \\
Max literals per clause ($L_{\max}$) & 4 \\
Spurious variables ($S$) & 3 \\
Spurious flip rate ($\rho$) & 0.3 \\
\bottomrule
\end{tabular}
\caption{Synthetic data generation parameters. All distributions are discrete uniform over the specified ranges.}
\label{tab:synthetic_params}
\end{table}

\paragraph{Scope of Coverage.} Our generator covers sparse DNF rules with bounded complexity ($K \leq 6$, $L \leq 4$) over moderate-sized variable sets ($N \leq 12$). This does not uniformly sample the space of all Boolean functions, many of which require exponentially many DNF clauses. Rather, it targets the sparse, interpretable rules that are the focus of rule induction research.

\section{Literal Feature Vector}
\label{sec:literal_features}

For each literal $l_j$ (where $j \in \{1, \ldots, 2N\}$ indexes both positive literals $x_i$ and negations $\neg x_i$), we compute a feature vector $\phi_j \in \mathbb{R}^{18}$ containing the following statistics computed from the episode $(X, Y)$:

\begin{table}[t]
\centering
\small
\begin{tabular}{lp{6cm}}
\toprule
\textbf{Feature} & \textbf{Description} \\
\midrule
$P(l_j{=}1 | y{=}1)$ & Truth rate among positive examples \\
$P(l_j{=}0 | y{=}1)$ & Complement of above \\
$\text{obs}^+$ & Observation rate among positive examples \\
$P(l_j{=}1 | y{=}0)$ & Truth rate among negative examples \\
$P(l_j{=}0 | y{=}0)$ & Complement of above \\
$\text{obs}^-$ & Observation rate among negative examples \\
$P(l_j{=}1)$ & Marginal truth rate \\
$P(l_j{=}0)$ & Complement of above \\
$\text{obs}$ & Overall observation rate \\
$\mathcal{H}(l_j)$ & Binary entropy: $-p\log p - (1{-}p)\log(1{-}p)$ \\
$\text{sgn}_j$ & Literal polarity: 1 for positive, 0 for negation \\
(reserved) & Zero-padded slot for future features \\
$\bar{c}_j$ & Mean absolute co-occurrence strength \\
$\bar{c}^+_{\text{abs}}$ & Mean $|$co-occurrence$|$ among positive examples \\
$\bar{c}^+$ & Mean co-occurrence among positive examples \\
$\bar{c}^-_{\text{abs}}$ & Mean $|$co-occurrence$|$ among negative examples \\
$\bar{c}^-$ & Mean co-occurrence among negative examples \\
$\Delta\bar{c}$ & Difference: $\bar{c}^+ - \bar{c}^-$ \\
\bottomrule
\end{tabular}
\caption{Components of the literal feature vector $\phi_j$. Co-occurrence is computed as the centered covariance between literal truth values.}
\label{tab:literal_features}
\end{table}

The co-occurrence strength captures how a literal's truth value correlates with other literals. For a literal $l_j$, we compute:
\begin{equation}
    c_{j,k} = \frac{1}{M} \sum_{m=1}^M (l_j^{(m)} - \bar{l}_j)(l_k^{(m)} - \bar{l}_k)
\end{equation}
where $\bar{l}_j = \frac{1}{M}\sum_m l_j^{(m)}$ is the mean truth value. The aggregate co-occurrence strength is $\bar{c}_j = \frac{1}{2N-1}\sum_{k \neq j} |c_{j,k}|$.

The class-specific co-occurrence features ($\bar{c}^+$, $\bar{c}^-$, etc.) are computed analogously but restricted to positive or negative examples respectively. These help identify literals that participate in conjunctive patterns within a class.

Observation rates handle missing data: when a literal's value is unknown for an example, that example is excluded from the truth rate calculation but contributes to the observation rate statistic. The observation-rate features ($\text{obs}^+$, $\text{obs}^-$, $\text{obs}$) therefore directly quantify the effective sample size used to estimate each truth rate, allowing downstream layers to discount statistically thin literals. This is the mechanism behind the missing-data behavior described in Section~\ref{sec:literal_stats}.

\section{Auxiliary Terms in $\mathcal{L}_{\text{cf}}$}
\label{sec:cf_extra_terms}

The deployed counterfactual objective adds two small-weight regularizers to $\mathcal{L}_{\text{nec}} + \mathcal{L}_{\text{spur}}$. Let $C_k^{(m)}$ be the clause truth value (Eq.~\ref{eq:clause_truth}), $\mathcal{P} = \{m : y_m = 1\}$ the set of positives, and $r_k^{(m)} = \text{softmax}_k(C_k^{(m)})$ the responsibility weights from the necessity term.

\textbf{Clause coverage overlap} ($\lambda_o = 0.1$) penalizes pairs of clauses that simultaneously fire on the same positive example, discouraging redundant coverage:
\begin{equation}
    \mathcal{L}_{\text{ovl}} = \frac{1}{|\mathcal{P}|\binom{K}{2}} \sum_{m \in \mathcal{P}} \sum_{k < k'} C_k^{(m)} C_{k'}^{(m)}
\end{equation}

\textbf{Counterfactual load balance} ($\lambda_c = 0.01$) is the negative mean per-positive responsibility entropy, encouraging each positive to be explained by more than one clause within the CF objective:
\begin{equation}
    \mathcal{L}_{\text{cf-bal}} = -\frac{1}{|\mathcal{P}|} \sum_{m \in \mathcal{P}} \left( -\sum_k r_k^{(m)} \log r_k^{(m)} \right)
\end{equation}

\section{Computational Scaling}
\label{sec:scaling_app}

\begin{figure}[t]
    \centering
    \includegraphics[width=0.9\columnwidth]{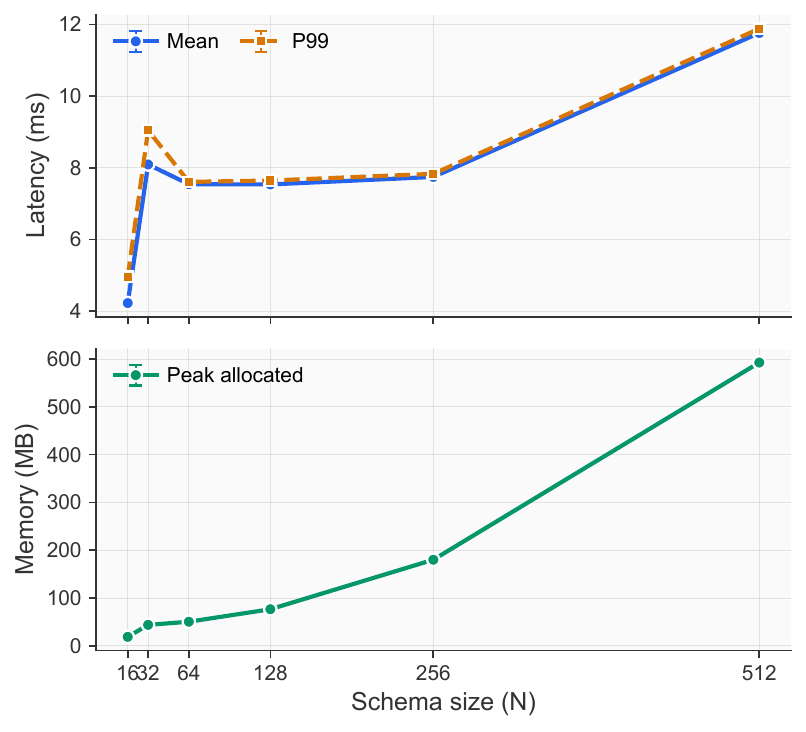}
    \caption{Computational scaling with problem size ($N$, left) and example count ($M$, right). Time remains nearly constant with $M$; memory scales quadratically with $N$ due to attention.}
    \label{fig:exp10a_scaling}
\end{figure}

Figure~\ref{fig:exp10a_scaling} reports inference latency and peak memory as the schema size $N$ and example count $M$ vary. Latency is nearly constant ($\sim$7.5ms) as $M$ increases from 32 to 512. For $N$-scaling, latency grows sub-linearly (4.2ms$\to$11.8ms for a $32\times$ increase from $N{=}16$ to $N{=}512$), while memory scales as $O(N^2)$ due to attention over $2N$ literals. At $N{=}512$, inference completes in under 12ms with 593MB peak memory.

\section{Baseline Descriptions}
\label{sec:baseline_descriptions}

This section describes the baseline methods compared against our Neural Rule Inducer (NRI). The baselines span gradient boosting methods (non-interpretable accuracy ceilings), generalized additive models, and interpretable rule-based classifiers. All baselines are trained per-dataset with 5-fold cross-validation.

\subsection{Non-Interpretable Ceilings}

These methods provide strong accuracy baselines but do not produce human-readable rules.

\paragraph{XGBoost.} XGBoost \cite{Chen2016XGBoost} is a highly optimized implementation of gradient boosted decision trees. It uses regularized learning objectives and efficient tree construction algorithms to achieve state-of-the-art performance on tabular data. The ensemble of trees is not directly interpretable, but provides a strong accuracy ceiling for comparison.

\paragraph{LightGBM (LGBM).} LightGBM \cite{Ke2017LightGBM} is a gradient boosting framework that uses histogram-based algorithms and leaf-wise tree growth for faster training and lower memory usage than traditional implementations. Like XGBoost, it serves as a non-interpretable accuracy ceiling.

\subsection{Generalized Additive Models}

Generalized additive models learn separate shape functions for each feature, providing partial interpretability through feature contribution graphs.

\paragraph{Explainable Boosting Machine (EBM).} EBM \cite{Lou2013EBM,Nori2019InterpretML} is a glassbox model that combines gradient boosting with generalized additive models. It learns a separate shape function for each feature, producing graphs showing how each feature contributes to predictions. While individual feature effects are interpretable, the model does not produce explicit logical rules and can include pairwise interaction terms that reduce transparency.

\subsection{Rule-Based Classifiers}

These methods produce explicit logical rules in various forms. Rule lists (RIPPER) evaluate rules sequentially; rule ensembles (RuleFit) weight rules linearly; tree-based methods (DT, FIGS) use hierarchical structures.

\paragraph{RIPPER.} RIPPER (Repeated Incremental Pruning to Produce Error Reduction) \cite{Cohen1995RIPPER} is a classic rule learning algorithm that grows rules greedily and then prunes them to optimize coverage and accuracy. It produces an ordered list of if-then rules that are evaluated sequentially. RIPPER is one of the most widely-used interpretable classifiers and serves as a primary baseline for rule induction.

\paragraph{RuleFit.} RuleFit \cite{Friedman2008RuleFit} extracts rules from an ensemble of decision trees and uses them as features in a sparse linear model. Each rule has an associated weight indicating its importance. While individual rules are interpretable, the weighted combination of many rules can reduce overall transparency compared to pure rule lists.

\paragraph{FIGS.} FIGS (Fast Interpretable Greedy-tree Sums) \cite{Tan2022FIGS} learns a sum of small decision trees, each constrained to be shallow. The algorithm uses greedy fitting with early stopping to prevent overfitting. The resulting model is interpretable as a collection of small trees whose outputs are summed.

\paragraph{Decision Tree (DT).} A standard CART-style decision tree \cite{Breiman1984CART} that recursively partitions the feature space using axis-aligned splits. We use scikit-learn's implementation with default hyperparameters. Decision trees are inherently interpretable as hierarchical rule structures, though the tree representation differs from DNF.

\paragraph{Decision Tree to DNF (DT-DNF).} A decision tree converted to Disjunctive Normal Form by extracting the conjunction of conditions along each path from root to a positive leaf. Each path becomes a clause in the resulting DNF rule. This provides a direct comparison of tree-derived DNF rules against our neural approach.

\subsection{Neural DNF Methods}

Neural approaches to DNF learning use differentiable approximations of logical operations, enabling gradient-based optimization of rule structures.

\paragraph{Neural DNF (Scratch) (N-DNF).} A neural network architecture designed to learn DNF rules, trained from scratch on each dataset. The architecture uses differentiable logic gates (sigmoid activations approximating AND/OR) similar to our approach, but without pre-training on synthetic data. This baseline tests whether per-dataset neural DNF learning outperforms our zero-shot transfer approach.

\subsection{Summary}

Table~\ref{tab:baseline_summary} summarizes the methods, their output types, and interpretability characteristics.

\begin{table}[h]
\centering
\small
\caption{Summary of baseline methods compared in our experiments.}
\label{tab:baseline_summary}
\begin{tabular}{llcc}
\toprule
Method & Type & Interp. & Training \\
\midrule
XGB & Gradient Boosting & No & Per-dataset \\
LGBM & Gradient Boosting & No & Per-dataset \\
EBM & GAM & Partial & Per-dataset \\
RIPPER & Rule List & Yes & Per-dataset \\
RuleFit & Rule Ensemble & Partial & Per-dataset \\
FIGS & Tree Sum & Yes & Per-dataset \\
DT & Decision Tree & Yes & Per-dataset \\
DT-DNF & DNF via Tree & Yes & Per-dataset \\
N-DNF & Neural DNF & Yes & Per-dataset \\
\bottomrule
\end{tabular}
\end{table}

\section{NRI Predicted Rules}
\label{sec:nri_rules}

This section presents example rules predicted by NRI on UCI datasets using 5\% training data with auto-tuned clause selection. Rules are shown after removing duplicate clauses. Logical operators: $\land$ (AND), $\lor$ (OR), $\lnot$ (NOT).

\begin{small}

\paragraph{adult} (1 clause)

\noindent ($\lnot$marital-status\_Never-married $\land$ $\lnot$relationship\_Not-in-family $\land$ $\lnot$relationship\_Own-child $\land$ $\lnot$sex\_Female)

\paragraph{breast-cancer-wisconsin} (1 clause)

\noindent (Clump\_Thickness\_gt\_median $\land$ Cell\_Size\_Uniformity\_gt\_median $\land$ Cell\_Shape\_Uniformity\_gt\_median $\land$ Bare\_Nuclei\_gt\_median)

\paragraph{car} (2+1+3+1 clauses for 4 classes)

\begin{itemize}[leftmargin=*,noitemsep]
\item \textbf{acc:} ($\lnot$persons\_2 $\land$ $\lnot$lug\_boot\_big $\land$ $\lnot$lug\_boot\_small $\land$ $\lnot$safety\_low) $\lor$ ($\lnot$persons\_2 $\land$ persons\_more $\land$ $\lnot$safety\_low)
\item \textbf{good:} ($\lnot$maint\_vhigh $\land$ $\lnot$persons\_2 $\land$ $\lnot$lug\_boot\_small $\land$ $\lnot$safety\_low)
\item \textbf{unacc:} (persons\_2 $\land$ safety\_low) $\lor$ (persons\_2 $\land$ $\lnot$safety\_high) $\lor$ (persons\_2)
\item \textbf{vgood:} ($\lnot$maint\_vhigh $\land$ $\lnot$persons\_2 $\land$ $\lnot$safety\_low $\land$ $\lnot$safety\_med)
\end{itemize}

\paragraph{credit-approval} (3 clauses)

\noindent ($\lnot$A11\_gt\_median $\land$ $\lnot$A15\_gt\_median $\land$ $\lnot$A7\_h $\land$ $\lnot$A10\_t) $\lor$ ($\lnot$A11\_gt\_median $\land$ $\lnot$A15\_gt\_median $\land$ $\lnot$A9\_t $\land$ $\lnot$A10\_t) $\lor$ ($\lnot$A11\_gt\_median $\land$ $\lnot$A15\_gt\_median $\land$ $\lnot$A6\_x $\land$ $\lnot$A10\_t)

\paragraph{diabetes} (1 clause)

\noindent (plas\_gt\_median $\land$ pres\_gt\_median $\land$ skin\_gt\_median $\land$ age\_gt\_median)

\paragraph{german-credit} (3 clauses)

\noindent (checking\_status\_no checking $\land$ $\lnot$purpose\_domestic appliance $\land$ $\lnot$employment\_1$\leq$X$<$4 $\land$ $\lnot$property\_magnitude\_no known property) $\lor$ (checking\_status\_no checking $\land$ $\lnot$employment\_1$\leq$X$<$4) $\lor$ (checking\_status\_no checking $\land$ $\lnot$purpose\_domestic appliance $\land$ $\lnot$employment\_1$\leq$X$<$4)

\paragraph{hepatitis} (3 clauses)

\noindent ($\lnot$BILIRUBIN\_gt\_median $\land$ $\lnot$ALK\_PHOSPHATE\_gt\_median $\land$ $\lnot$SGOT\_gt\_median) $\lor$ ($\lnot$BILIRUBIN\_gt\_median $\land$ $\lnot$ALK\_PHOSPHATE\_gt\_median $\land$ $\lnot$SGOT\_gt\_median $\land$ $\lnot$MALAISE\_yes) $\lor$ ($\lnot$BILIRUBIN\_gt\_median $\land$ MALAISE\_no $\land$ SPIDERS\_no $\land$ ASCITES\_no)

\paragraph{ionosphere} (2 clauses)

\noindent (a01 $\land$ a21\_gt\_median $\land$ a25\_gt\_median $\land$ a33\_gt\_median) $\lor$ (a01 $\land$ a21\_gt\_median $\land$ a25\_gt\_median)

\paragraph{kr-vs-kp} (6 clauses)

\noindent (bkxwp\_f $\land$ $\lnot$bkxwp\_t $\land$ bxqsq\_f $\land$ $\lnot$bxqsq\_t) $\lor$ ($\lnot$bkxwp\_t $\land$ bxqsq\_f $\land$ $\lnot$bxqsq\_t $\land$ $\lnot$wknck\_t) $\lor$ (bkxwp\_f $\land$ bxqsq\_f $\land$ $\lnot$bxqsq\_t $\land$ wknck\_f) $\lor$ ($\lnot$bkxwp\_t $\land$ $\lnot$bxqsq\_t $\land$ $\lnot$rimmx\_f $\land$ $\lnot$wknck\_t) $\lor$ (bkxwp\_f $\land$ bxqsq\_f $\land$ $\lnot$bxqsq\_t $\land$ wkna8\_f) $\lor$ (bxqsq\_f $\land$ $\lnot$bxqsq\_t $\land$ $\lnot$rimmx\_f)

\paragraph{mushroom} (5 clauses)

\noindent ($\lnot$odor\_n $\land$ $\lnot$gill-spacing\_w $\land$ $\lnot$stalk-root\_e $\land$ $\lnot$spore-print-color\_k) $\lor$ ($\lnot$odor\_n $\land$ $\lnot$stalk-root\_e $\land$ $\lnot$stalk-surface-above-ring\_s $\land$ $\lnot$spore-print-color\_k) $\lor$ ($\lnot$odor\_n $\land$ $\lnot$stalk-surface-above-ring\_s $\land$ $\lnot$spore-print-color\_k $\land$ $\lnot$spore-print-color\_n) $\lor$ ($\lnot$odor\_n $\land$ $\lnot$gill-spacing\_w $\land$ $\lnot$spore-print-color\_k $\land$ $\lnot$spore-print-color\_n) $\lor$ ($\lnot$odor\_n $\land$ $\lnot$stalk-surface-above-ring\_s $\land$ $\lnot$ring-type\_p $\land$ $\lnot$spore-print-color\_k)

\paragraph{nursery} (1+2+0+1+1 clauses for 5 classes)

\begin{itemize}[leftmargin=*,noitemsep]
\item \textbf{not\_recom:} (health\_not\_recom $\land$ $\lnot$health\_priority $\land$ $\lnot$health\_recommended)
\item \textbf{priority:} ($\lnot$parents\_great\_pret $\land$ $\lnot$has\_nurs\_very\_crit $\land$ $\lnot$health\_not\_recom) $\lor$ ($\lnot$parents\_great\_pret $\land$ $\lnot$has\_nurs\_very\_crit $\land$ $\lnot$health\_not\_recom $\land$ health\_recommended)
\item \textbf{recommend:} $\emptyset$ (empty rule)
\item \textbf{spec\_prior:} ($\lnot$parents\_usual $\land$ $\lnot$has\_nurs\_less\_proper $\land$ $\lnot$has\_nurs\_proper $\land$ $\lnot$health\_not\_recom)
\item \textbf{very\_recom:} ($\lnot$parents\_great\_pret $\land$ $\lnot$social\_problematic $\land$ $\lnot$health\_not\_recom $\land$ $\lnot$health\_priority)
\end{itemize}

\paragraph{spambase} (1 clause)

\noindent ($\lnot$word\_freq\_hp\_gt\_median $\land$ $\lnot$word\_freq\_hpl\_gt\_median $\land$ $\lnot$word\_freq\_george\_gt\_median $\land$ $\lnot$word\_freq\_meeting\_gt\_median)

\paragraph{tic-tac-toe} (1 clause)

\noindent ($\lnot$middle-left-square\_x $\land$ $\lnot$middle-middle-square\_o $\land$ middle-middle-square\_x)

\paragraph{vote} (1 clause)

\noindent ($\lnot$physician-fee-freeze\_n $\land$ $\lnot$education-spending\_n $\land$ $\lnot$crime\_n $\land$ $\lnot$duty-free-exports\_y)

\end{small}

\end{document}